\documentclass[letterpaper, 10 pt, conference]{ieeeconf}  

\IEEEoverridecommandlockouts                              

\usepackage{graphicx}
\usepackage{amsmath}
\usepackage{amssymb}
\usepackage{booktabs}
\usepackage{tabularx}
\usepackage{multirow}
\usepackage{placeins}    
\usepackage{array}
\usepackage[font=small]{caption}

\title{A Novel Reconfigurable Dexterous Hand Based on Triple-Symmetric Bricard Parallel Mechanism}

\author{
Chunxu Tian$^{1}$,
Zhichao Huang$^{1}$,
Hongzeng Li$^{1}$,
Bo Wang$^{1}$,
Jinghao Jia$^{1}$,
Yirui Sun$^{1}$,
and Dan Zhang$^{2,\dagger}$ 
\thanks{*This work was supported by the National Nature Science Foundation of China (grants 52305012).}%
\thanks{$^{1}$Institute of AI and Robotics, Academy for Engineering \& Technology, Fudan University, Shanghai 200433, China.}%
\thanks{$^{2}$Department of Mechanical Engineering, The Hong Kong Polytechnic University, Hung Hom, Hong Kong.}%
\thanks{$^{\dagger}$Corresponding authors: Dan Zhang (email: dan.zhang@polyu.edu.hk).}
}

\begin{document}

\maketitle
\thispagestyle{empty}
\pagestyle{empty}

\begin{abstract}
This paper introduces a novel design for a robotic hand based on parallel mechanisms. The proposed hand uses a triple-symmetric Bricard linkage as its reconfigurable palm, enhancing adaptability to objects of varying shapes and sizes. Through topological and dimensional synthesis, the mechanism achieves a well-balanced degree of freedom and link configuration suitable for reconfigurable palm motion, balancing dexterity, stability, and load capacity. Furthermore, kinematic analysis is performed using screw theory and closed-loop constraints, and performance is evaluated based on workspace, stiffness, and motion/force transmission efficiency. Finally, a prototype is developed and tested through a series of grasping experiments, demonstrating the ability to perform stable and efficient manipulation across a wide range of objects. The results validate the effectiveness of the design in improving grasping versatility and operational precision, offering a promising solution for advanced robotic manipulation tasks.
\end{abstract}

\section{Introduction}
\label{sec:introduction}

Dexterous robotic manipulators have emerged as a new class of robotic end-effectors with vast potential applications. Their multi-degree-of-freedom (DOF) structures, combined with multi-sensory capabilities, excellent adaptability, and diverse grasping modes, make them a key research focus in industrial robotics. Over the past few decades, the design of dexterous manipulators has progressed steadily \cite{lee2016kitech,martin2014modular}. Most dexterous grippers utilize anthropomorphic layouts where serial rotary joints are driven by tendons or cables, often incorporating springs or other compliant elements to achieve underactuated behavior \cite{dollar2010sdm,ma2014linkage,belzile2017optimal,chen2016adaptive}. However, the rigidity of conventional structures limits their adaptability to objects with varying geometry, scale, and surface properties.

Multi-fingered dexterous hands demonstrate excellent adaptability. Compared to series-linked structures, dexterous hands employing parallel mechanisms (PM) exhibit superior rigidity and stability, enabling higher precision and flexibility in complex grasping tasks while enhancing the manipulator's load-bearing capacity \cite{wang2024review}. These advantages enable parallel-structured dexterous hands to perform grasping and pinching tasks with rapid response and high positioning accuracy. In 2002, Lee and Tsai \cite{lee2002structural} introduced PM theory into the synthesis of dexterous hands, although their fingers remained serial. Under fingertip grasping, multi-DOF hands are kinematically equivalent to PMs---the object behaves as a moving platform and the fingers as kinematic limbs. This viewpoint subsequently promoted translational motion as an effective basis for high-precision, high-payload dexterous manipulators \cite{borras2015dimensional,ma2016spherical,borras2014analyzing}.

In recent years, Dai et al. \cite{dai2009orientation} designed a multi-finger dexterous hand with a reconfigurable palm. This design employs a reconfigurable mechanism as the palm, not only expanding the dexterous hand's workspace but also enhancing its dexterity, thereby offering a novel solution to the fixed-palm problem. However, current research on dexterous gripper palms primarily focuses on structural design and control optimization, often neglecting the role of the palm's own posture and shape during grasping. Consequently, some scholars have initiated studies on palms using reconfigurable mechanisms \cite{jin2020synthesis,tian2021design}. For examples, Patricia et al. \cite{capse2020exploring} designed a modular, 2-DOF deformable palm. Without adding extra motors, a soft cooperative drive mechanism synchronized palm and finger movements, improving the workspace and grasping adaptability. Lu et al.\ \cite{lu2021systematic} designed a reconfigurable underactuated dexterous hand with three underactuated fingers and a reconfigurable palm, enabling precise palm manipulation of diverse objects. Wei et al. \cite{an2021geometric} proposed a geometric design and dimensional synthesis method for a reconfigurable dexterous hand, investigating the impact of design parameters on the working space of the palm and fingers.

However, current research on dexterous robotic hands primarily focuses on structural design and control optimization, neglecting the role of the palm's own posture and shape during grasping. When grasping diverse objects in complex environments, factors such as palm posture, palm shape, and finger angles significantly impact grasping success rates. This research project aims to design a generalized parallel reconfigurable dexterous hand capable of multiple motion modes and adjustable palm posture. The structure is organized as follows: Section 2 establishes a topological analysis framework by discussing loop closure and contraction graphs (CGs). Section 3 performs kinematic analysis of the dexterous hand. Section 4 completes performance analysis, including workspace, Jacobian matrix, motion/force transmission analysis, and stability/stiffness analysis. Section 5 conducts experimental testing and analysis of the reconfigurable dexterous hand. Finally, Section 6 summarizes research findings and proposes future research directions.

\section{Graph Synthesis and Structural Design}

Following the design requirements identified in Section I, this section focuses on the structural synthesis of a closed-loop mechanism suitable for a reconfigurable palm.

\subsection{Loops and DOF of the Dexterous Hand}

For spatial PMs, the mobility can be calculated by the modified Grübler-Kutzbach formula~\cite{huang2011freedom}:
\begin{equation}
M = 6(n - g - 1) + \sum_{i=1}^{g} f_i + v - \zeta,
\end{equation}
where $M$ represents represents the DOF of the mechanism, $n$ is the total number of links, $g$ is the number of kinematic pairs, and $f_i$ is the DOF of the $i$-th pair. $v$ is the number of passive DOFs, and $\zeta$ is the redundant constraints.

To simplify topological synthesis, we introduce basic links, where $n_k$ is the count of basic links with $k$ kinematic pairs. The total links $n$ and kinematic pairs $g$ are then:
\begin{equation}
\begin{aligned}
n &= \sum_{k=2}^{m} n_k, \qquad
g = \sum_{i=1}^{g} f_i
   = \frac{1}{2}\sum_{k=2}^{m} k\,n_k .
\end{aligned}
\label{eq:basic_links}
\end{equation}

The number of independent loops $V$ of the mechanism can be calculated using Euler’s formula~\cite{hunt1978kinematic}:
\begin{equation}
\begin{aligned}
V &= g-n+1
   = \frac{1}{2}\bigl(n_3 + 2n_4 + 3n_5 + \cdots \bigr) + 1 .
\end{aligned}
\label{eq:V}
\end{equation}

Substituting the above equations yields the number of binary links $n_2$, which is a key parameter defining the mechanism topology:
\begin{equation}
\begin{aligned}
n_2 &= 5V - \sum_{k=3}^{n} n_k + \Phi,\\
\end{aligned}
\label{eq:n2}
\end{equation}
where $\Phi = M + \zeta - v - 1$. Compliant topological structures are synthesized by building CGs from basic link combinations, as listed in Table~\ref{tab:basic_links}.

\begin{table}[!t]
\centering
\setlength{\tabcolsep}{4pt}
\caption{Combinations of Basic Links for $V=1/2/3/4$.}
\label{tab:basic_links}

\begin{tabular*}{\columnwidth}{@{\extracolsep{\fill}}cccccc|cccccc@{}}
\toprule
$V$ & No. & $n_2$ & $n_3$ & $n_4$ & $n_5$ &
$V$ & No. & $n_2$ & $n_3$ & $n_4$ & $n_5$ \\
\midrule

1 & 1 & $5+\Phi$ & 0 & 0 & 0 &
1 & 9 & $14+\Phi$ & 6 & 0 & 0 \\
\midrule

2 & 2 & $8+\Phi$ & 2 & 0 & 0 &
2 & 10 & $15+\Phi$ & 4 & 1 & 0 \\
2 & 3 & $9+\Phi$ & 0 & 1 & 0 &
2 & 11 & $16+\Phi$ & 3 & 0 & 1 \\
2 & 4 & $11+\Phi$ & 4 & 0 & 0 &
2 & 12 & $16+\Phi$ & 2 & 2 & 0 \\
2 & 5 & $12+\Phi$ & 2 & 1 & 0 &
2 & 13 & $17+\Phi$ & 2 & 0 & 1 \\
2 & 6 & $13+\Phi$ & 1 & 0 & 1 &
3 & 14 & $17+\Phi$ & 1 & 1 & 0 \\
\midrule

3 & 7 & $13+\Phi$ & 0 & 2 & 0 &
3 & 15 & $17+\Phi$ & 0 & 3 & 0 \\
3 & 8 & $14+\Phi$ & 0 & 0 & 1 &
3 & 16 & $18+\Phi$ & 0 & 1 & 1 \\
\midrule

4 & -- & -- & -- & -- & -- &
4 & 17 & $18+\Phi$ & 0 & 0 & 2 \\
\bottomrule

\end{tabular*}
\end{table}

\FloatBarrier

\subsection{Topological Graph and Isomorphism Identification}

\begin{figure}[t]
\centering
\includegraphics[width=\linewidth]{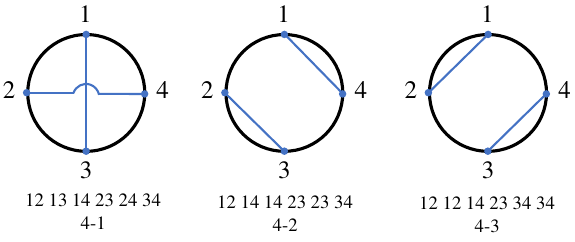}
\caption{Topological graphs and connection arrays.}
\label{fig:topo_graph}
\end{figure}

Although configurations with $V=2$ or $V \geq 4$ are theoretically feasible, $V=2$ provides insufficient structural complexity for spatial reconfiguration, while higher values of $V$ significantly increase mechanical complexity and reduce stiffness. Therefore, $V=3$ is selected as a compromise between reconfigurability, symmetry, and structural robustness.A topological graph uses nodes for joints and edges for links to show their connections. To achieve a balance of high flexibility and load capacity, the dexterous hand design requires a trade-off between stiffness, dexterity, and complexity. This design sets $V=3$ and uses configuration No.~4 from Table~\ref{tab:basic_links}, which has four ternary links (T) and $(11+\Phi)$ binary links. With four ternary links forming the core, there is only one circumferential arrangement ($TTTT$), resulting in three valid topological graphs~\cite{tian2019partially,tian2017structural}. Isomorphism identification is necessary to avoid redundancy from different graphical representations.By numbering the basic links, each connection is converted into a mathematical array, as shown in Fig.~\ref{fig:topo_graph}.

This study applies the Node Response Method (NRM) for analysis~\cite{xia2022novel}. The NRM verifies isomorphism by simulating layer-by-layer response transmissions and comparing output sequences from different initial nodes. This process yielded two non-isomorphic graphs with design potential: 4-1 and 4-2.

\begin{figure}[t]
\centering
\includegraphics[width=\linewidth]{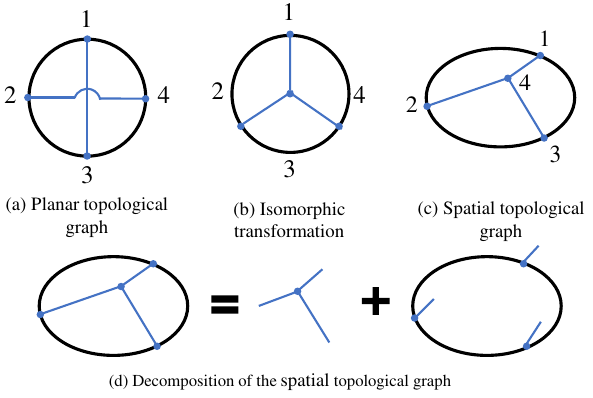}
\caption{Transformation and Decomposition of the Graph.}
\label{fig:graph_transform}
\end{figure}

Symmetrical structures are preferred in mechanism design for easier analysis, manufacturing, and maintenance. CG~4-1 has a triple-symmetrical spatial topology, with three identical branches connected to one vertex. This symmetry improves the load capacity and stability. Consequently, CG~4-1 was selected, with its various forms shown in Fig.~\ref{fig:graph_transform}.

\subsection{Structural Design Based on the Topological Graph}

Deconstructing the CG~4-1 spatial topology shows three branches, each with a ternary link (1, 2, or 3). One kinematic pair on each branch attaches to the base (4), while the other two connect the branches serially. Leveraging this triple symmetry, we designed a three-fingered symmetrical hand. This creates a six-bar closed-loop, with fingers on the binary links between ternary links 1, 2, and 3.

\begin{figure}[t]
\centering
\includegraphics[width=\linewidth]{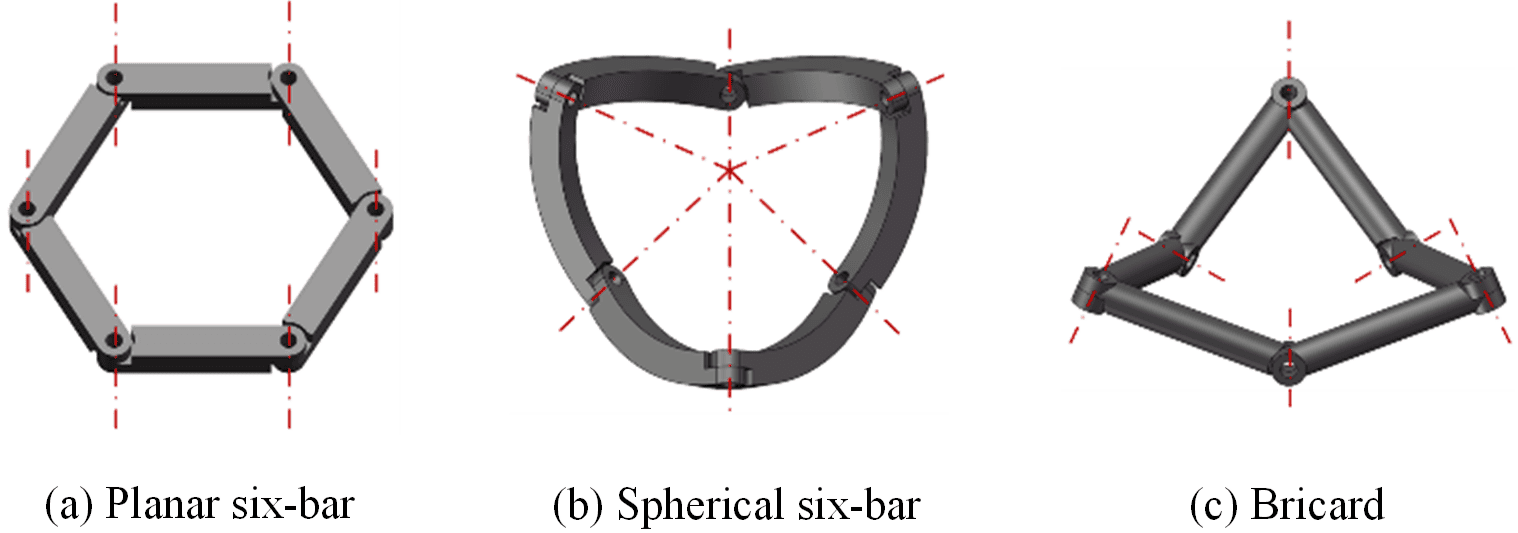}
\caption{Symmetrical six-bar mechanisms}
\label{fig:bricard}
\end{figure}

Among symmetrical six-bar options, the Bricard mechanism~\cite{chen2005threefold} was chosen for the morphing palm, as its spatial deployability offers superior palm flexibility and adaptability, as shown in Fig.~\ref{fig:bricard}. This structure offers a wider range of motion possibilities and expandability, making it widely applicable.

\begin{figure}[t]
\centering
\includegraphics[width=\linewidth]{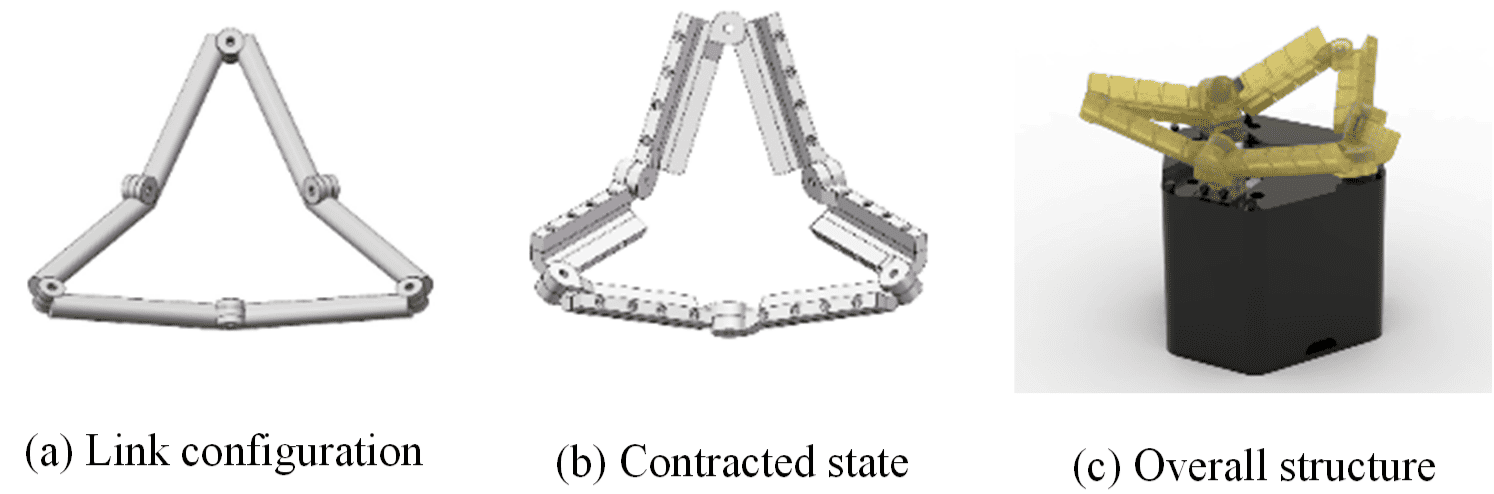}
\caption{Structural design of the triple-symmetric Bricard mechanism}
\label{fig:structure}
\end{figure}

After choosing a triple-symmetrical Bricard variant, two three-dimensional models were designed, as sketched in Fig.~\ref{fig:structure}. Model~4(a) is simple but less precise due to manufacturing variance. Model~4(b) is more complex and ensures higher precision via positioning holes, which adopts the precise Fig.~\ref{fig:structure}(b) configuration, with the final structure shown in Fig.~\ref{fig:structure}(c).

\section{Kinematic Analysis of the Mechanism}

\subsection{Structural Analysis of the Dexterous Hand}

\begin{figure}[t]
\centering
\includegraphics[width=0.9\linewidth]{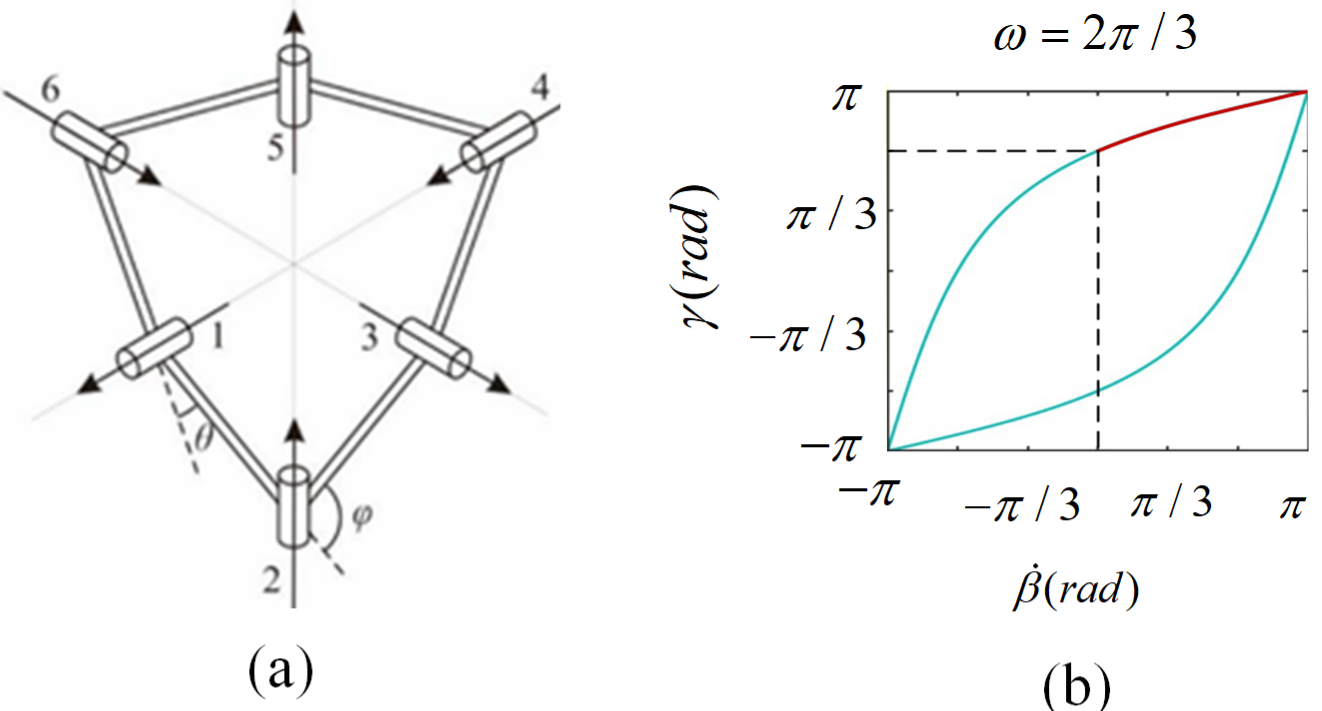}
\caption{(a) Triple-symmetric Bricard mechanism, where $\beta$ and $\gamma$ denote the rotation angles of adjacent revolute joints as defined in the local link frames; (b) corresponding curves of $\beta$ and $\gamma$ with $\omega = 2\pi/3$.}
\label{fig:bricard_geometry}
\end{figure}

Depending on the number and configuration of its links, the Bricard mechanism exhibits diverse structural characteristics. According to Denavit-Hartenberg (DH)~\cite{chen2005threefold}, the mobility of a single-loop mechanism is gained if and only if the ordered product of all homogeneous transformation matrices around the loop equal to the identity, namely,

\begin{equation}
\prod_{i=1}^{n} \mathbf{T}_{i(i+1)} = \mathbf{I},
\label{eq:DH}
\end{equation}
where $\mathbf{T}_{i(i+1)}$ denotes the transfer matrix between links. This mechanism features triple plane and rotational symmetry, with its geometric conditions shown in Fig.~\ref{fig:bricard_geometry}(a) and met as follows:
\begin{equation}
\left\{
\begin{aligned}
a_{12} &= a_{23} = a_{34} = a_{45} = a_{56} = a_{61} = l \\
\alpha_1 &= \alpha_3 = \alpha_5 = \omega \\
\alpha_2 &= \alpha_4 = \alpha_6 = 2\pi - \omega \\
\beta_1 &= \beta_3 = \beta_5 = \beta \\
\gamma_1 &= \gamma_3 = \gamma_5 = \gamma
\end{aligned}
\right.
\label{eq:geom_param}
\end{equation}
As shown in Fig.~\ref{fig:bricard_geometry}(a), the triple-symmetric Bricard mechanism possesses three symmetry planes, combining rotational and planar symmetries. Hence, it is adopted as the dexterous hand palm. Based on the DH closed-loop condition ~\eqref{eq:DH} and the geometric parameters ~\eqref{eq:geom_param}, the motion coordination equation of the mechanism is derived as

\begin{equation}
\begin{alignedat}{1}
&\sin^2\omega(\cos\beta+\cos\gamma)
+ (1+\cos^2\omega)\cos\beta\cos\gamma \\
&\quad + \cos^2\omega - 2\cos\omega\sin\beta\sin\gamma = 0
\end{alignedat}
\label{eq:7}
\end{equation}

In~\eqref{eq:7}, $\omega$ denotes the angle between adjacent joints,while $\beta$ and $\gamma$are link rotation variables. The continuous motion is ensured by setting $\omega = 2\pi/3$, maximizing workspace and avoiding interference.

Choosing $\beta$ as input, the feasible domain is restricted to  $\beta \in (0, \pi)$ and $\gamma \in (2\pi/3, \pi)$ to eliminate bifurcation ambiguity. Therefore, each $\beta$ corresponds to a unique $\gamma$, llustrated by the red curve in Fig.~\ref{fig:bricard_geometry}(b). the relationship is:
\begin{equation}
5 \cos \beta \cos \gamma
+ 4 \sin \beta \sin \gamma
+ 3 \cos \beta
+ 3 \cos \gamma
+ 1 = 0
\label{eq:8}
\end{equation}

Given $\omega$ and the admissible ranges of $\beta$ and $\gamma$, the schematic configuration of the triple-symmetric Bricard mechanism is illustrated in Fig.~6(a). In this mechanism, each pair of non-adjacent revolute joints intersects at a common point (indicated by the red lines), and the centers of these revolute joints form two parallel equilateral triangles (green lines). The line connecting the two intersection points is perpendicular to both triangular planes, which guarantees the geometric symmetry of the mechanism.

As depicted in Fig.~6(b), $M_i$ and $N_i$ $(i = 1, 2, 3)$ denote the centers of the revolute joints connecting the palm to the kinematic chains and fingers, respectively. Each link of the mechanism is represented by $M_iN_i$. Owing to the triple symmetry, $\triangle M_1M_2M_3$ and $\triangle N_1N_2N_3$ are parallel equilateral triangles, forming the fundamental geometric structure of the mechanism.

A reference frame $\{O_m\}: O_m - x_m y_m z_m$ is established on the plane of $\triangle M_1M_2M_3$, with the origin $O_m$ located at its centroid. The $x_m$-axis connects $O_m$ to $M_1$, the $z_m$-axis is perpendicular to the plane $\triangle M_1M_2M_3$, and the $y_m$-axis is determined by the right-hand rule. Point $G$ is defined as the centroid of $\triangle N_1N_2N_3$. The motion screws at $M_i$ and $N_i$ are denoted by $S_{M_i}$ and $S_{N_i}$, respectively, where $\beta$ is the input angle and $\gamma$ is the passive angle.

\begin{figure}[t]
\centering
\includegraphics[width=\linewidth,height=0.9\linewidth,keepaspectratio]{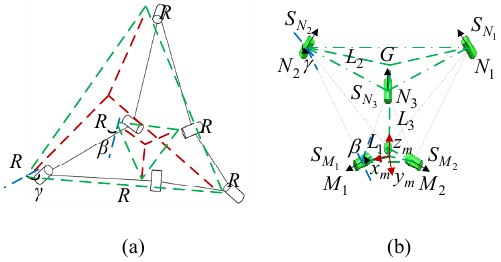}
\caption{(a) Schematic diagram of the triple-symmetric Bricard mechanism; (b) corresponding screw system representation.}
\label{fig:screw_system}
\end{figure}

The length of the Bricard is represented by $L$. For calculation convenience, let $O_mM_1 = O_mM_2 = O_mM_3 = L_1$, $GN_1 = GN_2 = GN_3 = L_2$ and $O_mG = L_3$. The parameters $L_1$, $L_2$ and $L_3$ can be expressed in terms of $L$, $\beta$, and $\gamma$.

\begin{equation}
\left\{\small
\begin{aligned}
L_1 &= \frac{2\sqrt{3}}{3} \, L \cos \frac{\gamma}{2}, \\
L_2 &= \frac{2\sqrt{3}}{3} \, L \cos \frac{\beta}{2}, \\
L_3 &= L \sqrt{
1 - \frac{4}{3}\cos^2 \frac{\gamma}{2}
  - \frac{4}{3}\cos^2 \frac{\beta}{2}
  + \frac{4}{3}\cos \frac{\gamma}{2}\cos \frac{\beta}{2}
}
\end{aligned}
\right.
\label{eq:9}
\end{equation}

\subsection{Forward kinematic analysis}

In the proposed dexterous hand model, we designate $L_1$ as the input variable and $L_3$ as the output variable to describe the position of the moving platform in the coordinate system, thereby establishing the forward kinematics. By combining equation (8), (9), and the double-angle formula, and eliminating $\beta$ and $\gamma$, the constraint equation between $L_1$ and $L_2$ is obtained as
\begin{equation}
45L_1^2L_2^2
- 12L^2(L_1^2+L_2^2)
+ 48L^2L_1L_2 \sqrt{A}\sqrt{B} = 0.
\label{eq:10}
\end{equation}

where
$A = 1-{3L_1^2}/{4L^2}$
and
$B = 1-{3L_2^2}/{4L^2}$
Based on equation (10), the constraint relationship between $L_1$ and $L_2$ can be visualized, as shown in Fig.~7(a). Since $\beta \in (0,\pi)$, $\gamma \in (2\pi/3,\pi)$, it follows that $L_1 \in (0,\sqrt{3}L/3)$, $L_2 \in (0,2\sqrt{3}L/3)$, which leads to the condition $L_1 < L_2$. Under this condition, the palm exhibits greater extension capability, resulting in an enlarged workspace for finger motion and thus enhanced dexterity of the hand. Consequently, the relationship between $L_1$ and $L_2$ is represented by the red curve in Fig.~7(a), indicating that within this interval, for any $L_2$, there exists a unique corresponding $L_1$.

Let $U = L_2^2 / L^2$, $V = L_1^2 / L^2$, move the root term and square it, and after organizing, simplify and obtain the quadratic equation about $V$ as shown as
\begin{equation}
A(U)V^2 + B(U)V + C(U) = 0 .
\label{eq:11}
\end{equation}

Where $A(U) = 16 + 72U + 81U^2$, $B(U) = -224U + 72U^2$, and $C(U) = 16U^2$. By applying the quadratic formula, the value of $L_2$ can be obtained. Therefore, the final expression is given by Eq.~(12):
\begin{equation}
L_2
=
L
\sqrt{
\frac{
112L_1^2L^2 - 36L_1^4
+ 32L_1^2L
\sqrt{12L^2 - 9L_1^2}
}{
16L^4 + 72L_1^2L^2 + 81L_1^4
}
}.
\label{eq:12}
\end{equation}

Therefore, given the value of $L_1$, $L_2$ can be uniquely determined from Eq.~(12), and subsequently $L_3$ can be uniquely obtained from Eq.~(13). In this way, the forward kinematics of the proposed mechanism is fully established.
\begin{equation}
L_3 = \sqrt{L^2 - (L_1^2 + L_2^2) + L_1 L_2}
\label{eq:13}
\end{equation}

\subsection{Inverse kinematics analysis}

In the inverse kinematics analysis, $L_3$ is selected as the independent variable, the transformation relationship of $L_1$ with respect to $L_3$ is derived. From Eq.~(13), it can be rearranged into a quadratic form in terms of $L_2$.

\begin{equation}
L_2^2 - L_1 L_2 + (L_1^2 + L_3^2 - L^2) = 0
\label{eq:14}
\end{equation}

Similarly, the solution $L_2$ can be derived:

\begin{equation}
L_2 = \frac{L_1 + \sqrt{4L^2 - 3L_1^2 - 4L_3^2}}{2}
\label{eq:15}
\end{equation}

By combining Eqs.~(12) and (15), a constraint equation depending only on $L_1$ and $L_3$ is obtained. Within the interval $L_1 \in (0,\sqrt{3}L/3)$ and $L_3 \in (0,L)$ for any given $L_1$, there exists a unique corresponding $L_3$. The functional relationship will be presented in Section~4.1 on workspace analysis. Hence, the inverse kinematics of the proposed mechanism is established.

\section{Performance Analysis of the Dexterous Hand}

Performance analysis supports the design and optimization of mechanical systems. Using the established kinematic models and constraints, the dexterous hand is evaluated in terms of workspace, motion/force transmission, stability, and stiffness to ensure accuracy and reliability for practical application.

\subsection{Workspace analysis}

Workspace characterizes the reachable positions of a mechanism and reflects its input--output capability. For the dexterous hand, inverse kinematics is solved under motion constraints and link parameters to obtain feasible inputs and determine the accessible workspace. The structural parameters are listed in Table~II.

\begin{table}[!t]
\centering
\caption{Structural Parameters of the Dexterous Hand}
\label{tab:II}
\setlength{\tabcolsep}{4pt}
\renewcommand{\arraystretch}{1.10}
\begin{tabular*}{\columnwidth}{@{\extracolsep{\fill}}lclc@{}}
\toprule
\textbf{Parameter} & \textbf{Value} & \textbf{Parameter} & \textbf{Value} \\
\midrule
$L$/mm   & 120         & $\omega$/rad & $2\pi/3$ \\
$L_1$/mm & $(0,70)$    & $\beta$/rad  & $(0,\pi)$ \\
$L_2$/mm & $(0,140)$   & $\gamma$/rad & $(2\pi/3,\pi)$ \\
$L_3$/mm & $(0,120)$   & --           & -- \\
\bottomrule
\end{tabular*}
\end{table}

\begin{figure}[t]
\centering
\includegraphics[width=\linewidth]{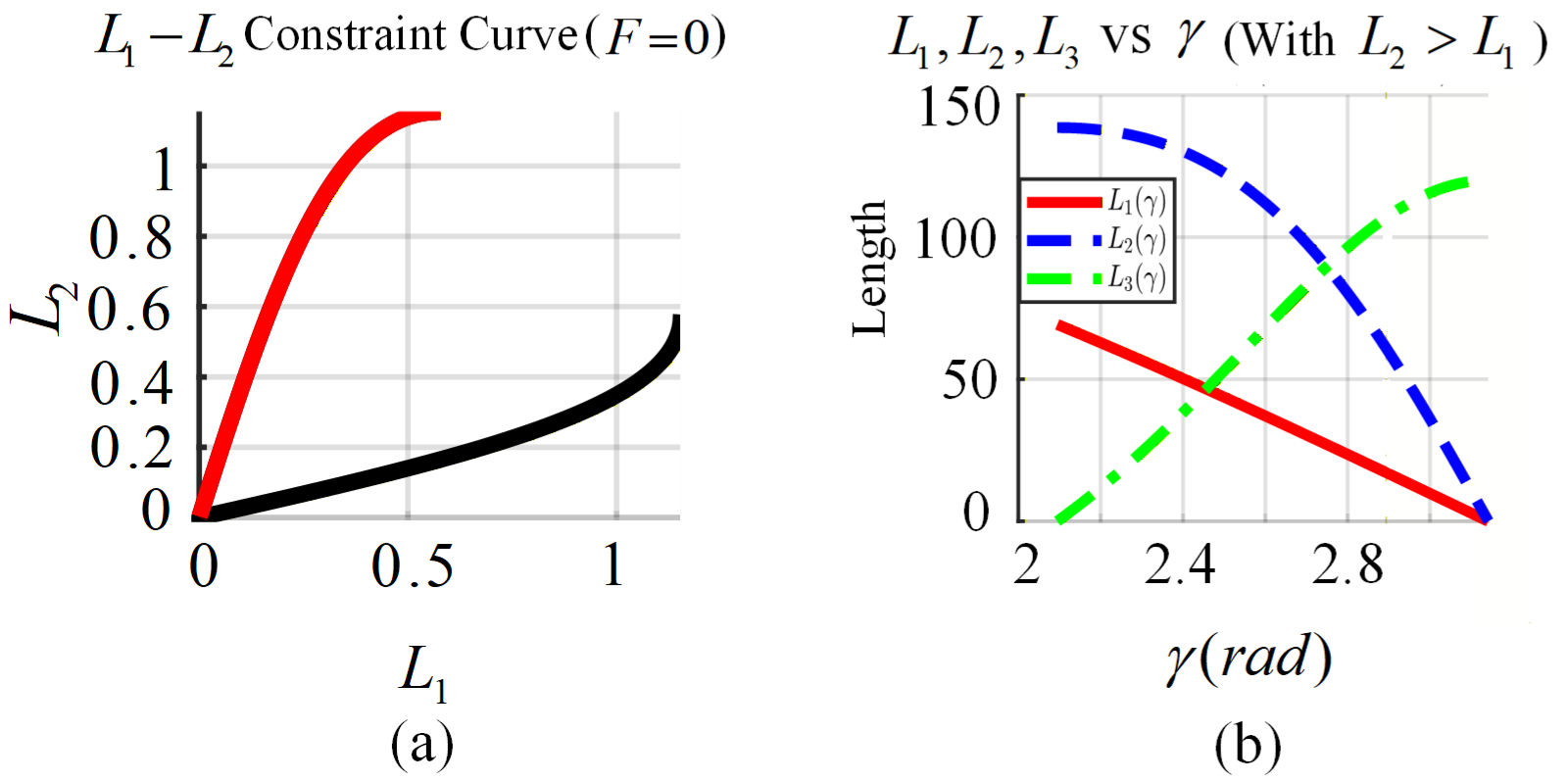}
\caption{(a) $L_1$ and $L_2$ (b) $L_1$, $L_2$ and $L_3$ with respect to $\gamma$.}
\label{fig:7}
\end{figure}

Considering the application scenarios of the dexterous hand, the following constraints are imposed to avoid mechanical interference during motion: $L_1 \in (0,\sqrt{3}L/3)$, $L_2 \in (0,2\sqrt{3}L/3)$ and $L_3 \in (0,L)$. Then, the reachable workspace of both the forward and inverse solutions is determined, along with the variation trends of $L_1$, $L_2$ and $L_3$ with respect to $\gamma$, as shown in Fig.~7(b).

\subsection{Jacobian Analysis}

The Jacobian maps joint velocities to end-effector velocities, enabling analysis of motion transmission, force distribution, singularities, and dexterity. For the dexterous hand, $J$ is obtained by differentiating the kinematic constraints with respect to the independent variables:

\begin{equation}
\dot{p} = J(q)\dot{q}
\label{eq:16}
\end{equation}

Here, $p \in \mathbb{R}^m$ denotes the Cartesian position of the end-effector (or a vector parameterizing both position and orientation), and $q \in \mathbb{R}^n$ represents the actuated joint variables. The vectors $\dot{p}$ and $\dot{q}$ denote the end-effector velocity and joint velocity, respectively. From Eqs.~(10) and (13), the relationship between the input variable $L_1$ and the output variable $L_3$ is obtained. Accordingly, the Jacobian matrix of the proposed mechanism can be expressed as:

\begin{equation}
\left\{
\begin{aligned}
J &= \frac{-2L_1 + L_2}{2L_3} - \frac{-2L_2 + L_1}{2L_3}
     \frac{N_1}{D_1}, \\[4pt]
N_1 &= 90L_1L_2^2 - 24L^2L_1
     + 48L^2L_2S_1S_2  \\
    &\quad - 36L_1^2L_2\frac{S_2}{S_1}, \\[4pt]
D_1 &= 90L_2L_1^2 - 24L^2L_2
     + 48L^2L_1S_1S_2  \\
    &\quad - 36L_2^2L_1\frac{S_1}{S_2}, \\[4pt]
S_1 &= \sqrt{1-\frac{3L_1^2}{4L^2}}, \\
S_2 &= \sqrt{1-\frac{3L_2^2}{4L^2}}.
\end{aligned}
\right.
\label{eq:17}
\end{equation}

Based on the geometric relationship in Eq.~(10), the Jacobian matrix can be numerically evaluated. Furthermore, the analytical expression of the Jacobian derived in Eq.~(17) enables the visualization of its distribution. The corresponding results are illustrated in Fig.~8(a).

\subsection{Stability and Stiffness Analysis}

Stiffness reflects a system's resistance to external disturbances, directly affecting stability and accuracy. High stiffness reduces deformation, while low or negative stiffness may cause buckling or oscillation. The proposed palm mechanism operates as a single-degree-of-freedom system during reconfiguration. For the single-degree-of-freedom reconfiguration motion, the Jacobian $J$ is a scalar given by $J = dL_3/dL_1$. Given the input stiffness $K_{L_1}$, the output stiffness $K_{L_3}$ is obtained via the stiffness transmission relation:
\begin{equation}
K_{L_3} = \frac{K_{L_1}}{J^2}
\label{eq:18}
\end{equation}

Stiffness describes how output displacement is amplified or attenuated relative to input via the Jacobian. A small Jacobian yields high stiffness, i.e., rigid response, while a large Jacobian reduces stiffness, i.e., compliant response. Thus, positive stiffness ensures stability, whereas near-zero or negative stiffness may induce instability. The variation of $K_{L_3}$ with respect to $L_1$ is shown in Fig.~8(b). The dexterous hand exhibits positive stiffness throughout its workspace, demonstrating stable kinematic behavior and effective resistance to external disturbances.

\begin{figure}[t]
\centering
\includegraphics[width=\linewidth]{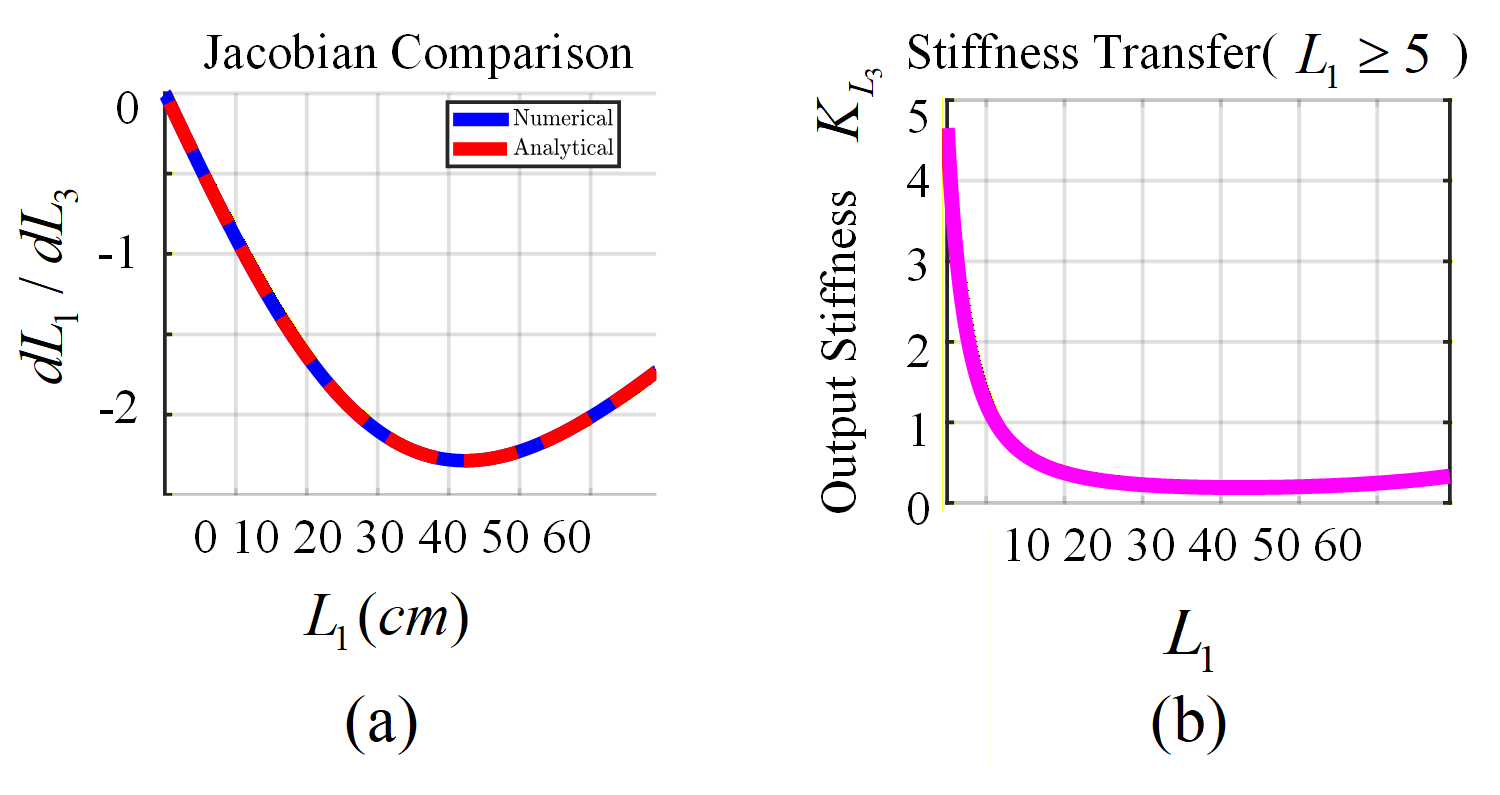}
\caption{(a) Jacobian Comparison (b) Relation between $K_{L_3}$ and $L_1$.}
\label{fig:8}
\end{figure}

\subsection{Motion/Force Transmission Analysis}

For non-redundant manipulators, the Local Transmission Index (LTI)~\cite{li2025optimal} can be used as a performance metric for motion/force transmission, which is defined as:
\begin{equation}
\kappa = \min\{\zeta_i,\sigma_i\}\,(i=1,2,\ldots,n)
\label{eq:19}
\end{equation}
where $\kappa$ represents the local transmission index, $\zeta_i$ refers to the input transmission efficiency of branch $i$, while $\sigma_i$ indicates the output transmission efficiency of branch $i$. Both values range within $(0,1)$, signifying the efficiency of energy transfer from the actuators to the branch and from the branch to the end-effector, respectively. A value approaching 1 suggests enhanced transmission performance. The formal definitions of these indices are as follows:
\begin{equation}
\left\{
\begin{aligned}
\zeta_i &= \left|\mathbf{S}_{T_i}\circ \mathbf{S}_{I_i}\right| \big/ \left|\mathbf{S}_{T_i}\circ \mathbf{S}_{I_i}\right|_{\max} \\
\sigma_i &= \left|\mathbf{S}_{T_i}\circ \mathbf{S}_{O_i}\right| \big/ \left|\mathbf{S}_{T_i}\circ \mathbf{S}_{O_i}\right|_{\max}
\end{aligned}
\right.
\label{eq:20}
\end{equation}

In Eq.~(16), $\mathbf{S}_{T_i}$ represents the transmission wrench screw (TWS) of branch $i$, $\mathbf{S}_{I_i}$ denotes the input twist screw (ITS), and $\mathbf{S}_{O_i}$ refers to the output twist screw (OTS). The transmission wrench screw is defined such that its reciprocal product with all passive joints in the branch equals zero, ensuring consistency with the kinematic constraints. The computation of the denominator's maximum value can be referred to~\cite{zhang2022indices}. In the reference coordinate frame $\{O_m\}: O_m-x_my_mz_m$, the coordinates of points $M_i$ and $N_i$ can be expressed explicitly. From these coordinates, the vectors corresponding to the revolute joints of each link can be obtained. By computing the cross products of these vectors, the unit vectors of the motion screws $s_i$, $i=M_1,M_2,M_3,N_1,N_2,N_3$ can be determined. Accordingly, the motion screw system of the triple-symmetric Bricard mechanism can be expressed as:

\begin{equation}
S_i = [s_i\ \ s_{0i}]^T = [s_i\ \ r_i \times s_i]^T
\label{eq:21}
\end{equation}

Finally, the motion screw system of the tri-symmetric Bricard mechanism can be expressed as:

Finally, the motion screw system of the tri-symmetric Bricard mechanism can be expressed as:

\begin{equation}
\setlength{\arraycolsep}{1.2pt}
\renewcommand{\arraystretch}{1.05}
\raisetag{16pt}
{\small
\begin{pmatrix}
S_{M_1}\\ S_{M_2}\\ S_{M_3}\\ S_{N_1}\\ S_{N_2}\\ S_{N_3}
\end{pmatrix}
=
\begin{pmatrix}
\frac{\sqrt{3}L_2L_3}{m_1} & 0 & m_3 & 0 & -2m_5 & 0 \\
-\frac{\sqrt{3}L_2L_3}{2m_1} & \frac{3L_2L_3}{2m_1} & m_3 & \sqrt{3}m_5 & m_5 & 0 \\
-\frac{\sqrt{3}L_2L_3}{2m_1} & -\frac{3L_2L_3}{2m_1} & m_3 & -\sqrt{3}m_5 & m_5 & 0 \\
\frac{\sqrt{3}L_1L_3}{m_2} & 0 & m_4 & 0 & -2m_6 & 0 \\
-\frac{\sqrt{3}L_1L_3}{2m_2} & \frac{3L_1L_3}{2m_2} & m_4 & \sqrt{3}m_6 & m_6 & 0 \\
-\frac{\sqrt{3}L_1L_3}{2m_2} & -\frac{3L_1L_3}{2m_2} & m_4 & -\sqrt{3}m_6 & m_6 & 0
\end{pmatrix}
}
\label{eq:22}
\end{equation}

Where, $m_1$-$m_6$ is a function of $L_1$, $L_2$ and $L_3$, with its explicit expression provided as follows:

\begin{equation}
\left\{
\begin{aligned}
m_1 &= \sqrt{3L_1^2L_2^2 + 3L_2^2L_3^2 - 3LL_2^3 + 3L_2^4/4} \\
m_2 &= \sqrt{3L_1^2L_2^2 + 3L_1^2L_3^2 - 3L_1^3L_2 + 3L_1^4/4} \\
m_3 &= \sqrt{3}L_1L_2/m_1 - \sqrt{3}L_2^2/(2m_1) \\
m_4 &= \sqrt{3}L_1L_2/m_2 - \sqrt{3}L_1^2/(2m_2) \\
m_5 &= L_1m_3/2 \\
m_6 &= -L_2m_4/2 - \sqrt{3}L_1L_3^2/(2m_2)
\end{aligned}
\right.
\label{eq:23}
\end{equation}

Through the analysis, it is found that the Bricard mechanism consists of six kinematic branches. Due to its triple-symmetric property, only two representative branches need to be analyzed. For convenience, Branches $M_1N_2$ and $M_1N_3$ are selected, where the input twist is aligned with the $x_m$-axis and the output twist is aligned with the $z_m$-axis, corresponding to linear velocities screw $S_i=(L_1\ 0\ 0;0\ 0\ 0)$, $S_o=(0\ 0\ L_3;0\ 0\ 0)$. Since branches $M_1N_2$ and $M_1N_3$ are symmetric with respect to the $x_mO_mz_m$-plane, only the transmission analysis of branch $M_1N_2$ is required. In this branch, $S_{M_1}$ denotes the driving joint screw and $S_{N_2}$ represents the passive joint screw, from which the transmission wrench screw (TWS) $S_{T_i}$ can be derived. Finally, the variation of $\zeta_i$, $\sigma_i$, $\kappa$ with respect to parameter $L_1$ is plotted, as shown in Fig.~9.
\begin{figure}[t]
\centering
\includegraphics[width=\linewidth]{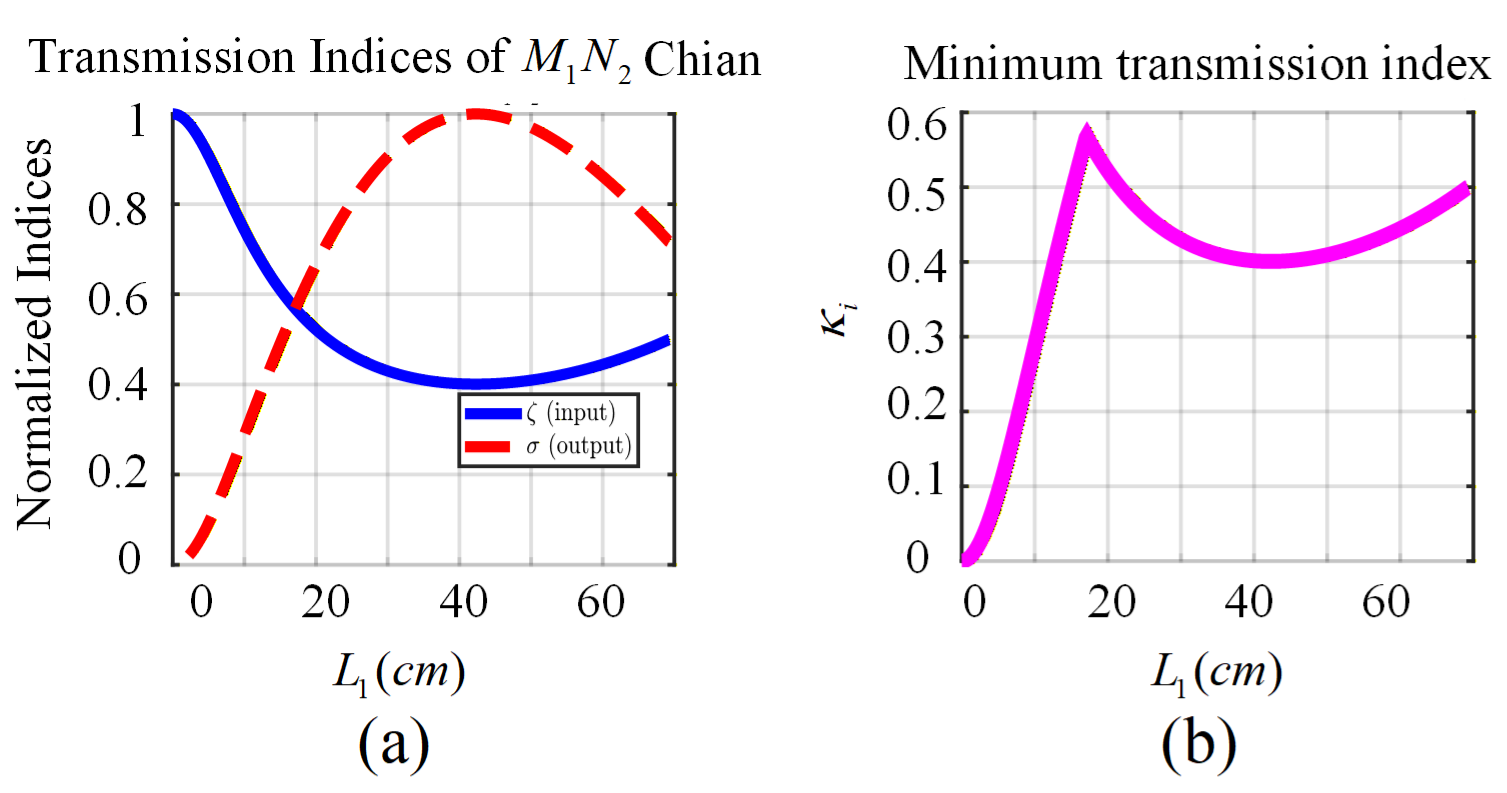}
\caption{The relationship between $\zeta_i$, $\sigma_i$, $\kappa$ and $L_1$.}
\label{fig:9}
\end{figure}

\section{Model Construction and Experiment}

Based on the preceding design and analysis, this section details the dexterous hand experiment through the prototype. It elaborates on the prototyping process and experimental validation for the universal parallel dexterous hand, systematically evaluating its actual performance.

\subsection{Design of the Dexterous Hand Model}

The dexterous hand employs an underactuated tendon-driven mechanism. Each finger consists of three sections---fingertip, middle phalanx, and base---connected by bearing-supported revolute joints. Finger flexion is achieved through tendon actuation, while torsion springs embedded in the joints provide passive restoring forces. The finger assemblies are mounted on the Bricard palm via mechanical connectors, with their motion constrained by the Bricard linkage.

To reduce friction during palm motion, the Bricard palm is connected to the base through a low-friction linear guide rail. The hand structure is fabricated from high-carbon steel, and all components undergo sandblasting and oxidation treatments to improve strength and wear resistance.

The hand is actuated by a cable-driven transmission system powered by six independent Feelech servo motors, each driving a single cable to avoid motion coupling. Three motors actuate the fingers, and the remaining three control palm opening and closing. All servos are controlled by an Arduino Nano with a URT-1 interface. Thin-film pressure sensors (RP-C7.6-ST) embedded at the fingertips provide real-time contact force feedback during grasping.
\begin{figure}[t]
\centering
\includegraphics[width=\linewidth]{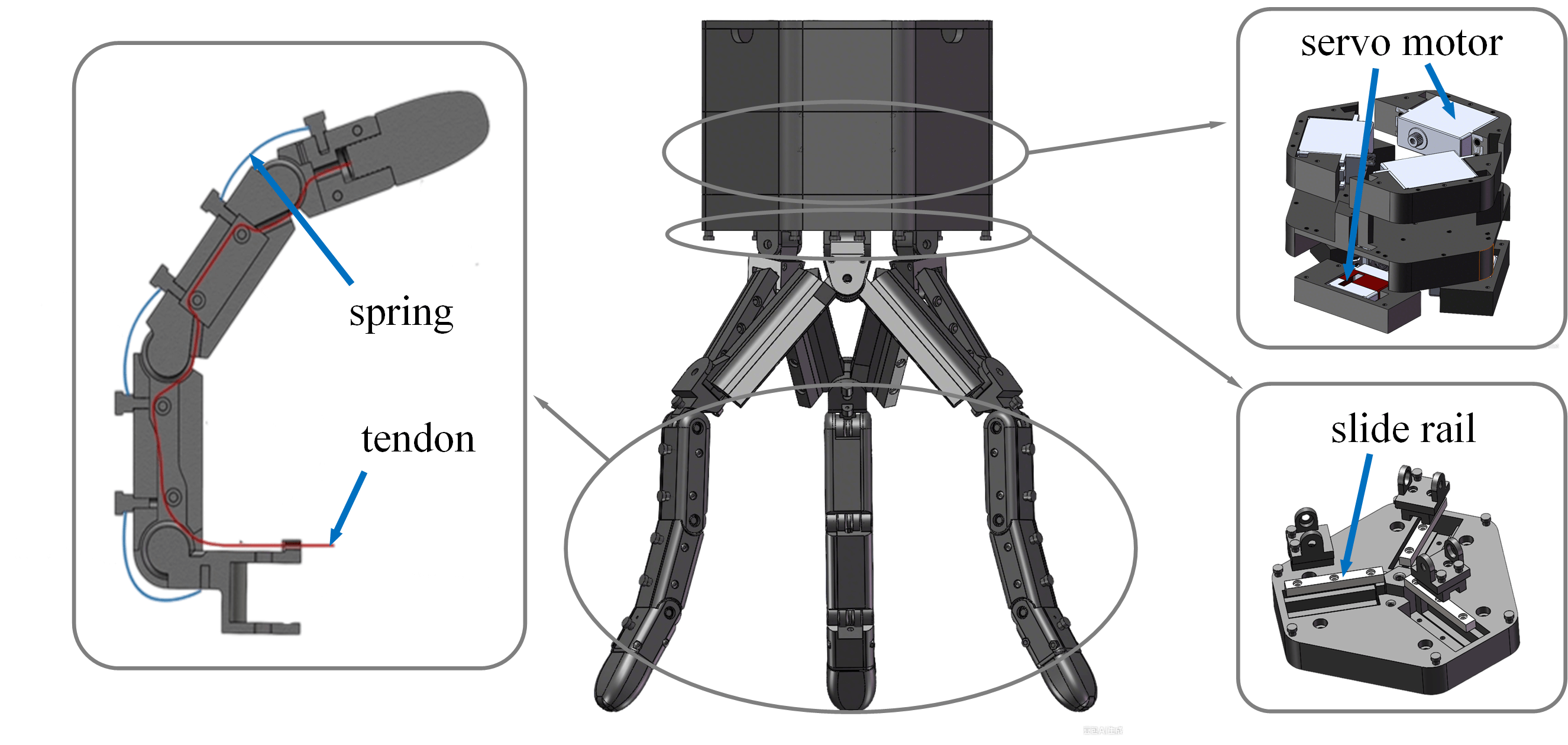}
\caption{Dexterous hand model structural overview.}
\label{fig:10}
\end{figure}
The gripping test validates the gripper's ability to grasp objects and highlights how its configurability expands the range of graspable items. The Bricard palm achieves different configurations by adjusting the clamping angle between linkages. In retracted mode, the end-effector size is minimized for grasping small objects in confined spaces. The semi-retracted mode suits medium-sized objects, while the fully extended palm provides a larger workspace for closed-grip manipulation of a broader variety of objects. The grasping performance is illustrated in Fig.~12.
\begin{figure}[t]
\centering
\includegraphics[width=\linewidth]{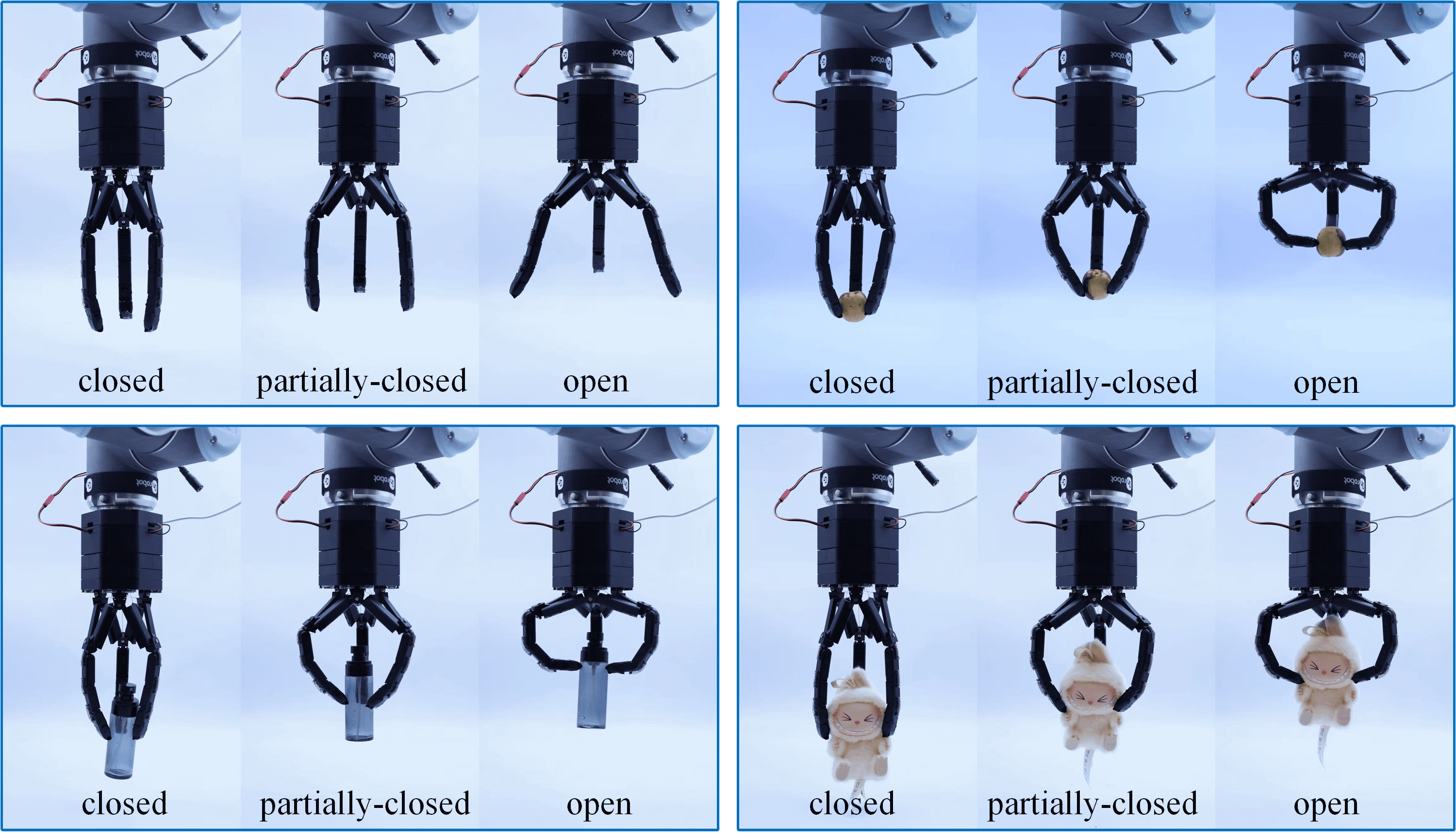}
\caption{Grasping comparison in different palm positions.}
\label{fig:12}
\end{figure}

\begin{table}[!t]
\centering
\caption{Grasping Results}
\label{tab:III}
\begin{tabular*}{\columnwidth}{@{\extracolsep{\fill}}lccc@{}}
\toprule
Objects & Close & Partially close & Open \\
\midrule
160mm & 0/3 & 0/3 & 3/3 \\
120mm & 0/3 & 3/3 & 3/3 \\
80mm  & 0/3 & 3/3 & 3/3 \\
40mm  & 3/3 & 3/3 & 3/3 \\
Mouse & 0/3 & 3/3 & 3/3 \\
Plates & 0/3 & 0/3 & 3/3 \\
Cylinders & 3/3 & 3/3 & 3/3 \\
Plush toys & 3/3 & 3/3 & 3/3 \\
Handles & 0/3 & 3/3 & 3/3 \\
Banana & 3/3 & 3/3 & 3/3 \\
Apple & 1/3 & 3/3 & 1/3 \\
Jujubes & 3/3 & 3/3 & 3/3 \\
Orange & 3/3 & 3/3 & 3/3 \\
Small bottles & 3/3 & 3/3 & 3/3 \\
Medium bottles & 3/3 & 3/3 & 3/3 \\
Large bottles & 3/3 & 1/3 & 1/3 \\
\bottomrule
\end{tabular*}
\end{table}

\subsection{Experiment and Grasping results}

To evaluate the grasping efficiency of finger structures across different object shapes, sizes, and weights, a series of standardized grasping tasks were designed. These validated the gripper's grasping capabilities, testing items ranging in size from 17\,mm to 180\,mm~\cite{wardcherrier2017modelfree}. The selected grasping test objects are shown in Fig.~11.
\begin{figure}[t]
\centering
\includegraphics[width=\linewidth]{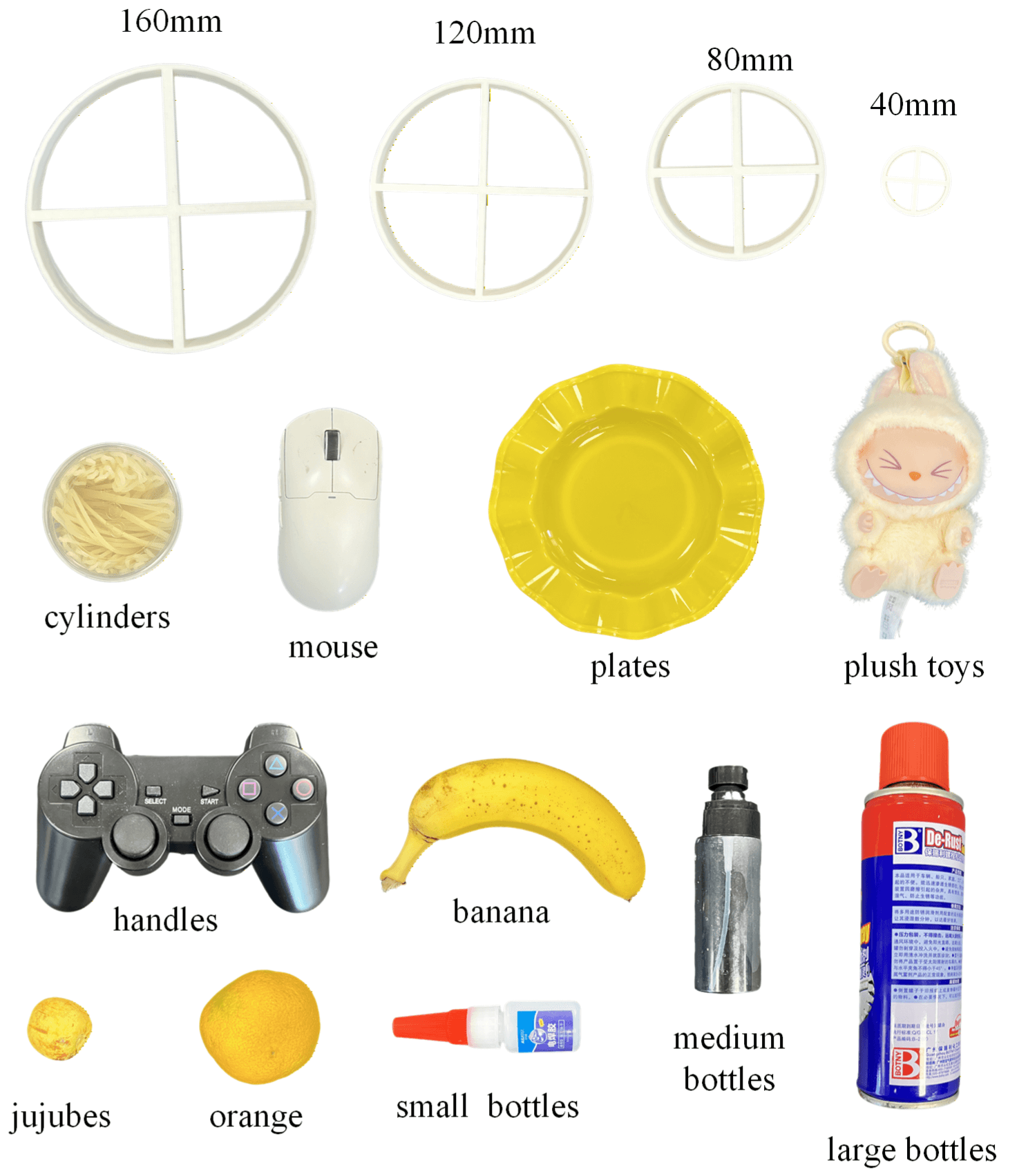}
\caption{Objects for evaluating gripper handling operations.}
\label{fig:11}
\end{figure}
Figure~13 illustrates some grasping results from the experimental process. First, the hand moves to the correct grasping position, and the fingers close to grasp the object. The gripper is then lifted 30 centimeters and held in that position for 3 seconds. Next, the gripper is controlled to move along the $x$-axis, followed by lowering the gripper to place the object down. If the object remains in a non-dropped state throughout the entire grasping process, the grasp is deemed successful~\cite{wang2020passively}.
\begin{figure}[t]
\centering
\includegraphics[width=\linewidth]{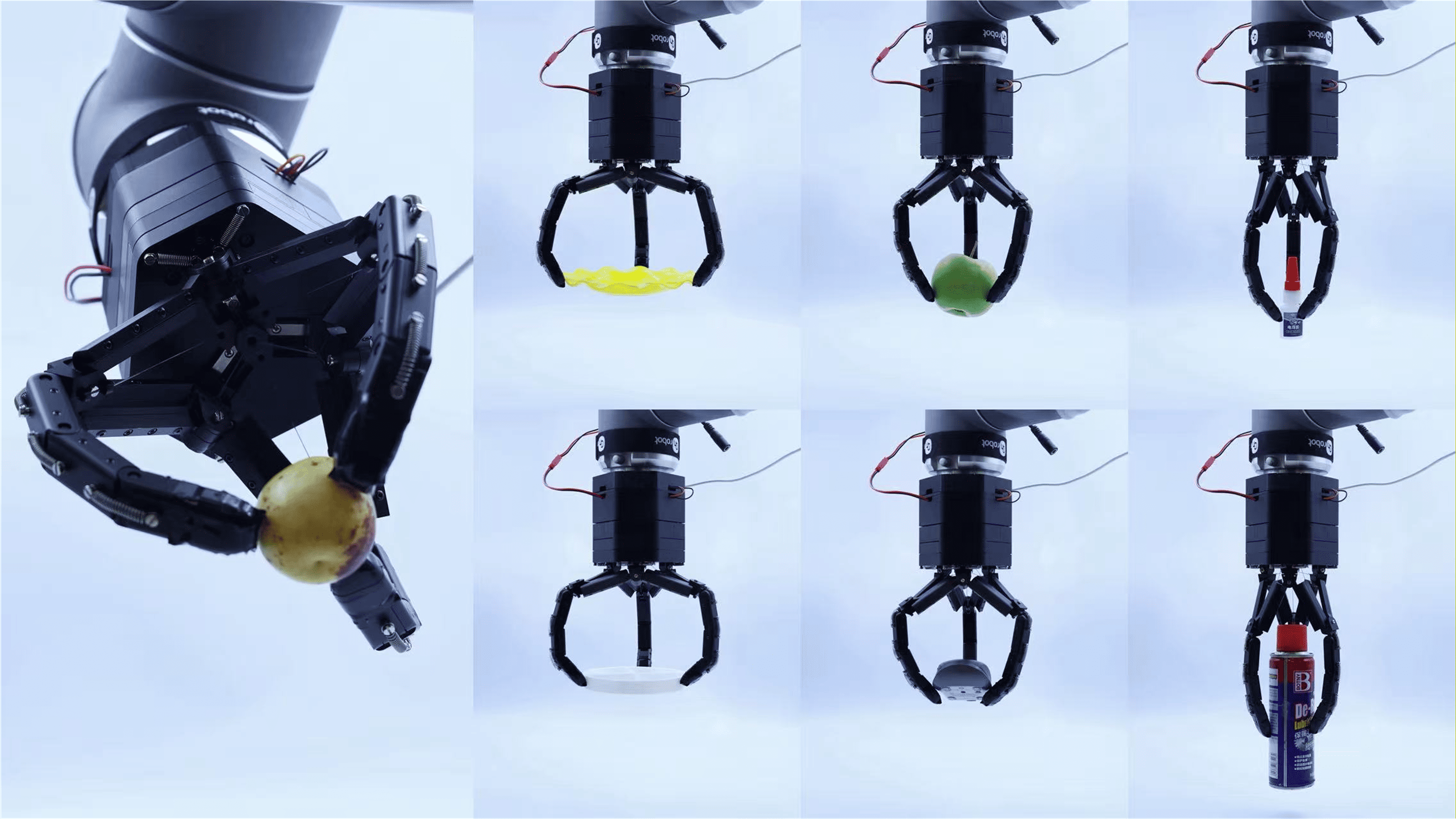}
\caption{Grasping results from the experimental process.}
\label{fig:13}
\end{figure}

To ensure grasping robustness, three grasping experiments were conducted for each object. Table~III documents grasping outcomes for various objects under different palm states. The numbers in the table represent the number of successful captures.

As shown in the grasping results, the dexterous hand achieves optimal grasping performance in the fully extended state, though its efficiency is lower for small yet heavy objects. The retracted state demonstrates superior grasping for small-volume objects and cylindrical items. The semi-retracted state exhibits distinct advantages when handling medium-to-small-volume objects with significant weight.

Furthermore, the dexterous hand can not only grasp objects horizontally but also manipulate objects in space. Fig.~13 illustrates some grasping results from the experimental process.

\begin{figure}[t]
\centering
\includegraphics[width=\linewidth]{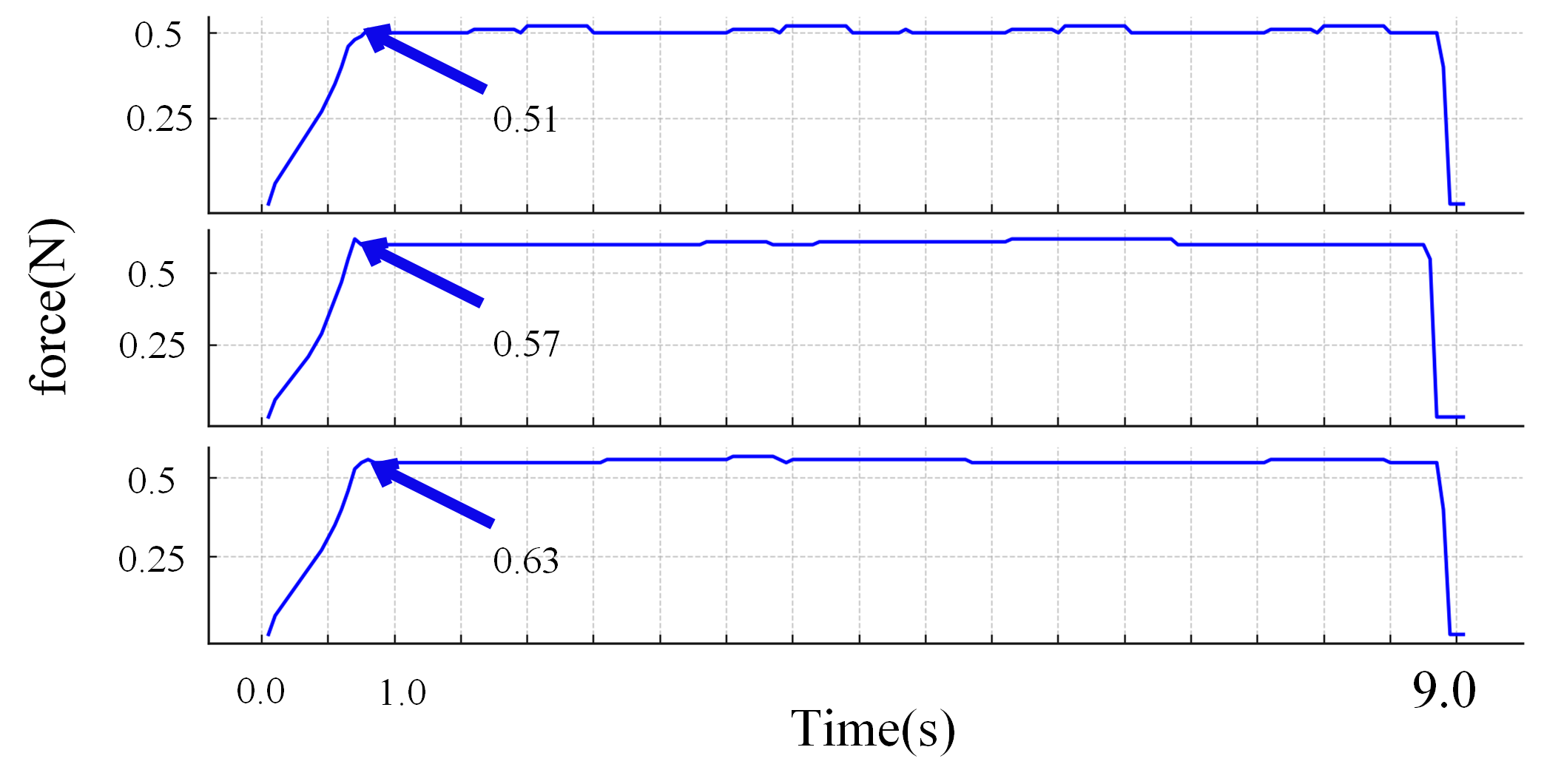}
\caption{The Pressure Sensor Data of The Capture Effect.}
\label{fig:14}
\end{figure}

The integration of end-effector pressure sensors enables the selection of the most suitable gripper configuration for different grasping tasks based on pressure readings. Fig.~14 illustrates the pressure sensor data recorded during the entire date palm grasping task performed with fingers. From top to bottom, the palm positions are fully closed, partially closed, and fully open. It is evident that during the task of grasping, the pressure sensor readings are lowest in the fully closed position, indicating the lowest energy consumption. This preliminary finding suggests that adopting a fully closed palm position is the optimal choice for small object grasping tasks.

\section{CONCLUSIONS}

In this study, a novel reconfigurable dexterous hand based on a triple-symmetric Bricard mechanism has been proposed, designed, and validated. The integration of a reconfigurable parallel palm structure significantly enhances the hand’s adaptability and workspace, allowing it to accommodate a diverse range of object geometries. Comprehensive kinematic and performance analyses confirm the motion characteristics, force transmission efficiency, and structural stability. Experimental results further demonstrate the prototype’s effectiveness in executing reliable grasps under various configurations. Future work will focus on optimizing actuator integration, improving real-time control strategies, and extending the design to multi-finger configurations for more complex manipulation tasks.


\bibliographystyle{IEEEtran}
\bibliography{references}

\end{document}